# A Self-rescue Mechanism for an In-pipe Robot for Large Obstacle Negotiation in Water Distribution Systems

Saber Kazeminasab
*Department of Electrical and Computer Engineering*
*Texas A&M University*
College Station, TX, USA
skazeminasab@tamu.edu

Moein Razavi
*Department of Computer Science and Engineering*
*Texas A&M University*
College Station, TX, USA
moeinrazavi@tamu.edu

Sajad Dehghani
*Department of Mechanical Engineering*
*University of Techran*
Tehran, Iran
sajjadreali@aut.ac.ir

Morteza Khosrotabar
*Deparments of Mechanical Engineering*
*University of Tehran*
Tehran, Iran
mkhosrotabar@ut.ac.ir

Katherine Banks
*College of Engineering*
*Texas A&M University*
College Station, TX, USA
k-banks@tamu.edu

*Abstract:* Water distribution systems (WDS) carry potable water with millions of miles of pipelines and deliver purified water to residential areas. The incidents in the WDS cause leak and water loss, which imposes pressure gradient and public health crisis. Hence, utility managers need to assess the condition of pipelines periodically and localize the leak location (in case it is reported). In our previous works, we designed and developed a size-adaptable modular in-pipe robot [1] and controlled its motion in in-service WDS. However, due to the linearization of the dynamical equations of the robot, the stabilizer controller which is a linear quadratic regulator (LQR) cannot stabilize the large deviations of the stabilizing states due to the presence of obstacles that fails the robot during operation. To this aim, we design a "self-rescue" mechanism for the robot in which three auxiliary gear-motors retract and extend the arm modules with the designed controller towards a reliable motion in the negotiation of large obstacles and non-straight configurations. Simulation results show that the proposed mechanism along with the motion controller enables the robot to have an improved motion in pipelines.

*Keyboards: In-pipe Robot, Self-rescue Mechanism, Wall-press Mechanism, Stabilizer Controller.*

## I. Introduction

Pipeline networks are one of the strategic infrastructures that carry oil, gas, and potable water in millions of miles of lines around the globe. The aging pipelines are prone to corrosion, damages, and even incidents in the network cause leak and water loss. The water loss phenomenon wastes a great deal of purified water that costs much for the authorities, and the amount associated with water loss is not negligible. For example in the US and Canada, water loss is responsible for 15%-25% and 20% waste of fresh water, respectively [2], [3]. Hence, it is required to assess the condition of pipelines periodically, and in case of a leak, the utilities need to localize it [4]. To this aim, mobile sensors are suggested that go inside pipes and move with water flow. These mobile sensors move in-pipes passively and perform the desired task during operation. However, since they are passive, the operator may lose them in the pipelines [5] and they cannot inspect long distances of pipelines. To address passiveness, in-pipe robots are designed in which their motion is active with actuators that move them in pipes. The operator can control the motion of these in-pipe robots to be independent of flow. The operation condition of in-pipe robots are pipelines with varying sizes and a high-pressure and velocity flow is present in which execute many disturbances on the robot during operation. Besides, pipelines' maps are not accurate [6] and due to sediments in pipelines, the circumferential geometry of pipes is not known and there are many uncertainties in pipelines that require the robotic systems to be stable and robust to these disturbances and uncertainties [7]. The aforementioned conditions are related to straight paths in pipelines. Pipelines comprise different non-straight configurations like bends, tees and in-pipe robots need to pass through. However, it is not possible to control the motion of the robots with one control algorithm, considering the conditions in the pipelines and the different geometry and configurations. Hence, there is a need for multi-phase motion control that enables the robot to move in a straight path and steer the robot in non-straight configurations of pipelines. In our previous works, we designed an under-actuated modular self-powered robot that can adapt to varying pipe sizes [1] and characterized it to have fully automotive motion in in-service distribution networks [8]. We also designed a multi-phase motion control algorithm for our robot that enables reliable motion in different configurations of pipelines [9]. The robot switches between different phases of the controller based on the configuration of the pipeline. In the straight paths, a stabilizer and velocity tracking controller controls the motion of the robot. In bends and T-junctions, another phase of the multi-phase controller steers the robot in the desired direction.

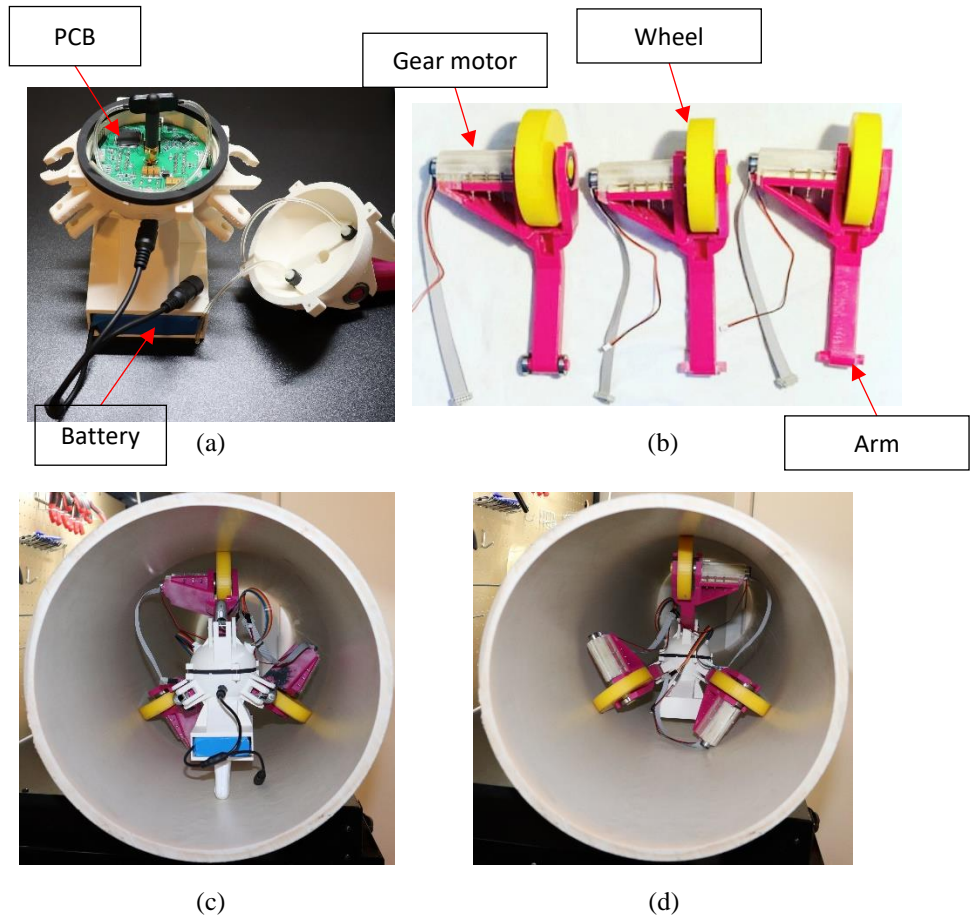

Fig. 1. Our Proposed In-pipe Robot. (a) The Central Processor and Its Different Components. (b) The Arm and Actuator Modules. (c) Front View of the Prototyped Robot. (d) Rear View of the Prototyped Robot.

Another challenge with the in-pipe robots that are self-powered is wireless communication [10]. There is high signal attenuation in the environment of water, soil, and rock that makes wireless communication a challenging task [11]. We designed a wireless robotic system in our previous work and proposed an operation procedure for the robot in [12] that facilitates smart navigation and data transmission during operation through wireless control. We also developed a localization and navigation for the robot based on the particle filtering method multi-phase motion control algorithm [13], [14] that removes the need for wireless communication. However, there is still a challenge with the motion of the robot; the stabilizer controller that is based linear quadratic regulator (LQR) is designed based on the linearized dynamic equations [15] is not able to stabilize the states of the robot in the case of large deviations due to large obstacles and non-straight configurations.

Hence, failure is probable for the robot during operation in the large deviations of the stabilizing states. In this work, we design a self-rescue mechanism that enables the robot to negotiate large deviations of stabilizing states and control its motion based on the feedback from the main actuators.

The remainder of the paper is organized as follows. In section II, the robotic system and the self-rescue mechanism is described. In section III, a static force analysis is done and the dynamical behavior of the robot is presented. The paper is then concluded in section V.

## II. In-pipe Robot Design

The robot comprises a central processor that hosts the sensor modules for parameter measurement and electronic components. Three adjustable arm modules are connected to the central processor with 120° angle and make the outer diameter of the robot to be adjustable to pipe diameters range from 9in-22in. Each arm module includes an arm and a passive spring that is connected to the arm in a way that one end is connected to the arm and the other end is connected to the central processor. At the end of each arm module, an actuator module is connected that moves the robot inside pipe. Each actuator module includes gear motor that is connected directly to a wheel, and a pair of ball bearings that connects the wheel on the arm and facilitates friction-free rotation about the arm. A battery is located in the central processor and provides power for all parts of the robot during operation. Fig. 1 shows the overall design of the proposed robot and its different components.

## III. Self-rescue Mechanism

In this section, we propose the self-rescue mechanism. First, we develop an argument for the operation situation of the robot and calculate the required torque and force for the mechanism. Then, we explain how we designed the mechanism.

### A) Static Force Analysis

The goal of self-rescue mechanism design is to activate the motion of the arms around their rotational joint ($O$ in Fig. 2) and enable them to rotate in the desired direction by a desired angle. For each arm, the passive spring facilitates the counter-clockwise motion for the arm and a torque is needed to facilitate the clockwise motion. To this aim, we considered a torque around $O$ in Fig. 2 and need to calculate it in all arm configurations. However, this torque to retract the arm is variable in each configuration of the arm since in each arm configuration, the passive spring has a specific deformation. To ensure the contact between the wheels and the pipe in severe cases based on the pipe's radius and geometry of the obstacle, we need to define the maximum torque needed to be exerted on the robot's arms. Based on the assumption that the robot moves quite slowly inside the pipe, the spring force can be obtained via the static force analysis; by doing torque equilibrium around point $O$ in Fig. 2, we have:

$$F_{spring}[t \sin(90 - \theta)] = (F_N - W_{arm})a \cos(\beta) \quad (1)$$

$$F_{spring} = \frac{1}{t \cos(\theta)}(F_N - W_{arm})a \cos(\beta) \quad (2)$$

where $F_N$ is the normal force exerted on the wheels from pipe wall, $W_{arm}$ is the arm's weight, $t$ is the length between points $A$ and $O$ (13 mm), $a$ is the length between points $O$ and $B$ (82 mm), $\theta$ is the angle between the vertical axis and spring line of action and $\beta$ is the angle between the arm and the horizontal axis (Fig. 2). Besides, we can compute the spring force based on the force-displacement equation for linear springs. We assume that at $\theta = 0°$, spring is at its rest length (Fig. 2).

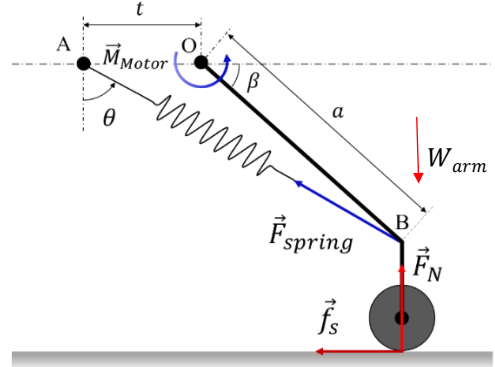

Fig. 2. Free body diagram of the arm module in the robot. $M_{Motor}$ is the torque applied to the arms to pass the obstacles in front of the robot.

$$F_{spring} = k[\sqrt{(t + a \cos(\beta))^2 + (a \sin(\beta))^2} - \sqrt{a^2 - t^2}] \quad (3)$$

By setting (2) and (3), we can find $F_N$ in terms of spring's stiffness and $\theta$ angle.

$$F_N = k[\sqrt{(t + a \cos(\beta))^2 + (a \sin(\beta))^2} - \sqrt{a^2 - t^2}]\frac{t \cos(\theta)}{a \cos(\beta)} + W_{arm} \quad (4)$$

Substituting different values for the spring's stiffness $k$ in (4) gives different normal forces, $F_N$ in terms of the arm angle $\theta$. Positive values for $F_N$ denote that there is a contact between the wheels and the pipe. The robot has no relative displacement in the vertical axis with respect to the pipe, so there is a force equilibrium condition along the vertical axis (Fig. 3).

$$F_N = F'_N + W_{total} \quad (5)$$

Based on (5), value of $F_N$ should be greater than $W_{total}$, to ensure that contact of the upper arm is in contact with the pipe. By considering this constraint on $F_N$ and based on (4), there is a specified interval for the angle $\theta$ which guarantees the contact of the upper arm with the pipe wall. If we increase the stiffness, the interval would be wider (Fig. 4). There is a trade-off between wider size-availability and manufacturing process in which higher stiffness results in wider range of $\theta$ but it makes the manufacturing process difficult (i.e. the spring anchors on the arms and the central processor needs to be sufficiently strong). On the other hand, lower stiffness makes the installation more facile but decrease the range of $\theta$ angle. We chose the stiffness to be 3 N/mm which results the range of angle $\theta$ to be between 20° and 66°. The required torque for the arm motors is obtained by maximizing the moment of the spring force around the arm's joint that results in around 2200 N.mm.

Hence, we analyzed the spring mechanism and selected a value for the spring stiffness in a way that prevents loss of traction between the wheels and the pipe wall. We also calculated the maximum value for the torque that is needed to retract the arm associated with the spring in this geometry.

*B) Self-rescue Mechanism Design*

In the previous section, we calculated the maximum required torque for the motor in point $O$ in Fig. 2 to retract the arm its range of motion. However, in order to provide that torque, we have space limitation in the robot preventing reliable assembling. Moreover, since the power for the parts is supplied by battery, we should consider power supply as a factor that affects the selection of motors and gearheads. We chose three EC 20 flat Ø20 mm, brushless, 5 W motors, each with a nominal torque of 8.58 N.mm and each motor is geared with GP 22 C Ø22 mm, 0.5 - 2.0 N.m planetary gearhead from Maxon Motors Inc©, with a reduction ratio of 270:1 resulting in the maximum torque capacity of 2316.6 N.mm. In order to install the gear motors on the robot, instead of attaching the gear motors into point $O$ in Fig. 2, we designed three waterproof platforms on the central processor in which each platform is under the dorsal side of the arms and locate the gear motors. In this platform, the gear motor is located in a slotted base and a motor cover is located above the gear motor and fixes it to the base with some pairs of screw-nuts. The slots in the motor cover are filled with moldable glue that prevents idle rotation of the gear motor in the base and also isolates it electrically (see Fig. 5a). Also, an anchor is created on the dorsal side of each arm and a flexible monofilament string connects the arm to the gear motors on the platform and transfer the power from the gear motors to the arm (see Fig. 5b).

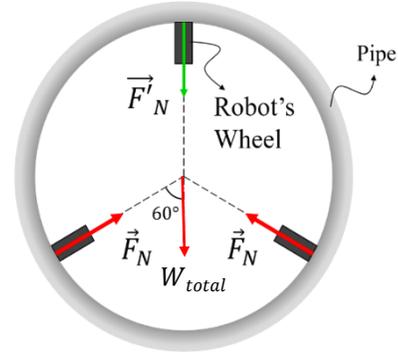

Fig. 3. View of the wheels and cross-section of the pipe. Here $F_N$ is the normal force on the lower arms and $F'_N$ is the normal force exerted on the upper arm.

In this section, we designed the concept of a self-rescue mechanism and characterized it based on the geometry of the robot and operation condition. We also, designed the mechanism in SolidWorks and addressed the challenges associated with the design. In the next section, we provide the dynamical modeling of the self-rescue mechanism.

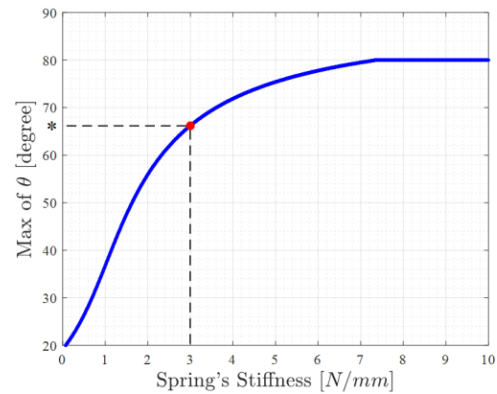

Fig. 4. Maximum allowable value for $\theta$ angle in each spring's stiffness. To have a contact between the upper arm and the pipe, $F'_N$ has to be positive. By substituting different values for $k$ in (4), we can find permissible range for $\theta$ that results in positive values for $F'_N$.

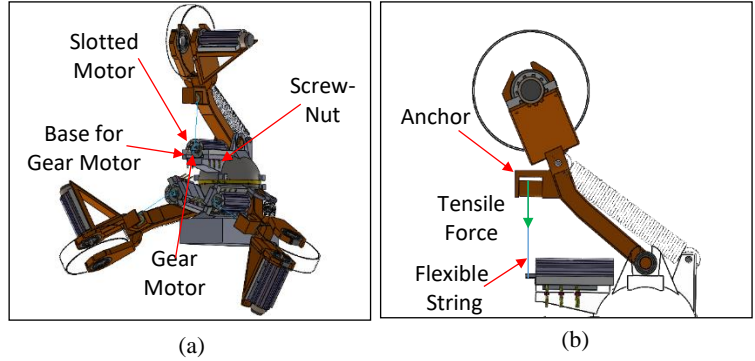

Fig. 5. (a) Platforms for the Gear Motors in the Self-rescue Mechanism. b) The Anchor on the Dorsal Side of the Arm and the Flexible String. The flexible string transfers the gear motors to the arm and has tensile force during operation.

## IV. Self-rescue Dynamical Modeling

This robot is classified as under-actuated mechanical system in which the number of control inputs is less than the system's degrees of freedom. It also works in an extremely uncertain and chattered environment. The nonlinearly in the system and the operating condition requires us to have an exact mathematical modelling of the robot. Due to the importance and complexity of the problem (robot dynamics modeling), the process of extracting robot dynamic equations is devided into two parts. First, the equations of the robot are extracted without the self-rescue mechanism, and in the next step, the equations are generalized to the desired model which includes the self-rescue mechanism.

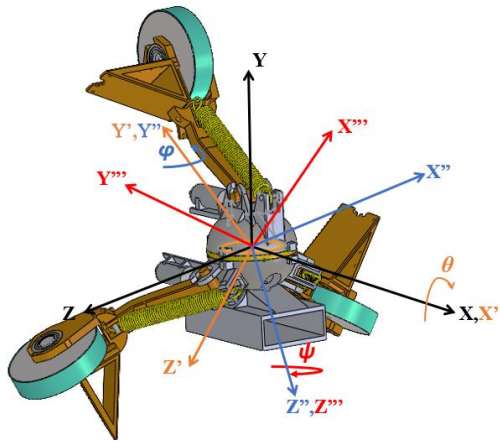

Fig. 4. Robot and Rotated Coordinates.

### A) Dynamical Modeling without Self-Rescue Mechanism

For the robot without self-rescue mechanism, we have derived the dynamic equation by conventional methods, in this paper the Lagrange method. As we know, this method is complex and computationally expensive, while giving us the compact dynamic model of system. Lagrangian is defined as:

$$\mathcal{L} = T - U \qquad (6)$$

Where $T$ and $U$ are the systems kinetic and potential energy, respectively. To extract these parameters, the robotic kinematic information (position and speed of links) is required, which is explained in [1], and the robot model is reviewed, and here again, it is briefly explained.

This robot has six degrees of freedom (DoF), Hence we have considered six generalized coordinates (GC) for this robot; three translations along x, y and z and three rotations $\theta$, $\phi$, and $\psi$ about x, y and z axes, respectively. We have considered these rotations ($\theta$, $\phi$ and $\psi$) as roll, pitch, and yaw, respectively.

For these six GCs, we have derived generalized velocity (GV) ($\boldsymbol{V}$ and $\boldsymbol{\omega}$) in the following; Eq. 7 shows the linear velocity of the robot.

$$\boldsymbol{V} = \begin{bmatrix} \dot{x} \\ \dot{y} \\ \dot{z} \end{bmatrix} \qquad (7)$$

As mentioned before, we have considered Euler 1-2-3 (roll, pitch and yaw) for rotational movement. So, for this choice, the first rotation ($\theta$) has transfers the XYZ coordinates to X'Y'Z' coordinates. The second and third rotations ($\phi$ and $\psi$), transfer X'Y'Z' and X"Y"Z" to X"Y"Z" and X'''Y'''Z''', respectively. (See Fig. 4). According to Euler 1-2-3, the angular velocity is expressed as:

$$\boldsymbol{\omega} = \dot{\theta} e_X + \dot{\phi} e_{Y'} + \dot{\psi} e_{Z''} \qquad (8)$$

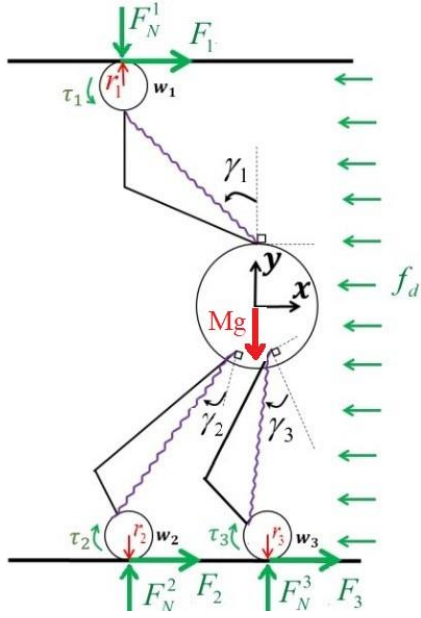

Fig. 5. The external forces from the fluid and pipe to the robot. a) $f_d$ is drag force, b) $F_i$ is friction force, and c) $F_N^i$ is normal force.

In Eq. 8, $e_X$, $e_{Y'}$, and $e_{Z''}$ are the unit vectors of X, Y' and Z" axes. By Euler transformation matrices, the expression of these vectors in the final coordinate system would be:

$$e_X = \cos(\phi)\cos(\psi) e_{X'''} - \cos(\phi)\sin(\psi) e_{Y'''} + \sin(\phi) e_{Z'''} \quad (9)$$

$$e_{Y'} = \sin(\psi) e_{X'''} + \cos(\psi) e_{Y'''} \quad (10)$$

$$e_{Z''} = e_{Z'''} \quad (11)$$

Plugging Eqs. (9)-(11) in Eq. (8) results in:

$$\boldsymbol{\omega} = \begin{bmatrix} \dot{\phi}\sin(\psi) + \dot{\theta}\cos(\phi)\cos(\psi) \\ \dot{\phi}\sin(\psi) - \dot{\theta}\cos(\phi)\sin(\psi) \\ \dot{\psi} + \dot{\theta}\sin(\phi) \end{bmatrix} \quad (12)$$

Now that we have the linear and angular velocities of the robot, the kinetic energy of the robot would be:

$$T = \tfrac{1}{2}\omega^T I_G \omega + \tfrac{1}{2} V^T M V \quad (13)$$

Where $M$ and $I_G$ are the robot's mass and mass moment of inertia matrices. For calculation of $I_G$, we decompose the robot to simple components (which th integral has been simplified) such as two hemispheres, three arms and one box. Then the moment of inertial can be expressed as:

$$I_G = \sum \left[ R_i^T I_{G_i} R_i + \int (d_i^2 I + d_i d_i^T) dm \right] \quad (14)$$

Where $I_{G_i}$ is the moment of inertia of the component $i$ in its principal coordinate, $R_i$ is the transformation matrix which it transforms the principal coordinates to the coordinates that are parallel to main coordinates, $d_i$ is the vector that transforms the body coordinates to the global coordinates and $I$ is a 3×3 eye matrix. For simple shapes, the integration is simplified in the form of Eq. (15):

$$\int (d_i^2 I + d_i d_i^T) dm = m_i (d_i^2 1 + d_i d_i^T) \quad (15)$$

Where $m_i$ is the mass of component $i$. So far, we calculated the kinetic energy, $T$, in Eq. (6). Another parameter in this equation is the potential energy ($U$). For deriving the potential energy, we have only considered the robot's weight as the potential term. Thus, we have:

$$U = mgy_G \quad (16)$$

Where $y_G$ is the displacement along y axis, which is the second term of the product $T_{4\times4} \times \rho_{G_0}$. $\rho_{G_0}$ is the position vector of the robot's center of mass (CoM) and $T_{4\times4}$ is the transformation matrix presented as:

$$T_{4\times4} = \begin{bmatrix} c\psi c\phi & s\psi c\theta + c\psi s\phi s\theta & s\psi s\theta - c\psi s\phi c\theta & x \\ -s\psi c\phi & c\psi c\theta - s\psi s\phi s\theta & c\psi s\theta + s\psi s\phi c\theta & y \\ s\phi & -c\phi s\theta & c\phi c\theta & z \\ 0 & 0 & 0 & 1 \end{bmatrix} \quad (17)$$

where $s$ and $c$ denote for sine and cosine functions, respectively. After calculating the kinetic and potential energies, we plug the Lagrangian in Eq. (18) as:

$$\frac{d}{dt}\left(\frac{\partial \mathcal{L}}{\partial \dot{q}_i}\right) - \frac{\partial \mathcal{L}}{\partial q_i} = Q_{nc_i} \quad (18)$$

Where $q$ is GC of the robot and $Q_{nc_i}$ is generalized force (GF). We have calculated $Q_{nc_i}$ by virtual work method. To calculate the $Q_{nc_i}$, first, the external forces acting on the robot must be determined. As shown in Fig. 5, $\boldsymbol{f}_d$, $F_i$, and $F_N^i$ are the external forces; $\boldsymbol{f}_d$ is the drag force, $F_i$, and $F_N^i$ are friction and normal forces applied on each wheel, respectively. For this problem, the forces must be written in local coordinates. Therefore we have:

$$\boldsymbol{f}_d = F_D R \begin{bmatrix} 1 \\ 0 \\ 0 \end{bmatrix} \quad (19)$$

$$F_N^i = F_s^i \boldsymbol{R} \begin{bmatrix} 0 \\ \cos(\mu_i) \\ \sin(\mu_i) \end{bmatrix} \quad (20)$$

where $\mu_i = \frac{2\pi}{3}(i-1)$, and $i = 1,2,3$.

$$F_i = r\tau_i \boldsymbol{R} \begin{bmatrix} 1 \\ 0 \\ 0 \end{bmatrix} \quad (21)$$

where $F_D$, $F_s^i$ are the value of drag force (which is applied to the robot due to moving in flow. This force is measured by Fluent software), and spring force applied to arm $i$, respectively. **R** is the rotation matrix that is calculated as:

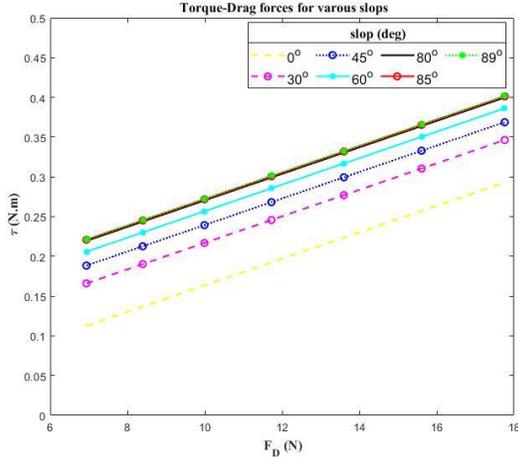

Fig. 6. Investigating the simultaneous effect of drag force change due to robot speed change and pipe slope on motor torques

$$R = \begin{bmatrix} c\psi c\phi & s\psi c\theta + c\psi s\phi s\theta & s\psi s\theta - c\psi s\phi c\theta \\ -s\psi c\phi & c\psi c\theta - s\psi s\phi s\theta & c\psi s\theta + s\psi s\phi c\theta \\ s\phi & -c\phi s\theta & c\phi c\theta \end{bmatrix} \quad (22)$$

Now we write the virtual work relation for these forces as follows:

$$\sum (Q_{nc_i} \delta q_i) = \sum (F_N^i . \delta \rho_i) + \sum (F_i . \delta \rho_i) + f_d . \delta \rho_G \quad (23)$$

The coefficient of each GC gives the GF.

*B) Dynamical Modeling with Self-rescue Mechanism*

In the previous subsection, we derived the dynamical equations of the robot without the self-rescue mechanism. In this subsection, we add the mechanism and revise the equations. By adding the desired mechanism, the only difference in the new system with the previous one is in the normal force, which is rewritten as follows:

$$F_N^i = (F_s^i - T^i) R \begin{bmatrix} 0 \\ \cos(\mu_i) \\ \sin(\mu_i) \end{bmatrix} \quad (24)$$

where $T^i$ is the tension force by the self-rescue mechanism.

*C) Functionality Evaluation of Equations*

So far, we have derived the equation of the motion of the robot by the Lagrange method and simplification assumptions. In this part, we validate the equations and also provide an estimate for the amount of torque that the gear motors need to provide in different operation conditions. In our previous work [13], we calculated the maximum of drag force that is applied on the robot to be 6 N, based on computational fluid dynamics (CFD) work. In addition, we considered the inclination angles of the pipe during motion (0°-90°). Fig. 6 shows the required torques on the motors in different drag forces and inclination angles of the pipe. The drag force in Fig. 8 varies from 6.3 N (when the robot is stopped) up to 18 N (when moving with 0.3 m/s velocity) and we have:

1. When the robot is stopped in a 0° slope, the static motor's torque is 0.11 N.m.
2. When the robot is stopped in a 90° slope, the static motor's torque is 0.22 N.m.
3. When the robot moves with 0.3 m/s velocity in a straight path with 0° slope, the torque that each motor can withstand is 0.3 N.m.
4. When the robot moves with 0.3 m/s velocity in a straight path with 90° slope, the torque that each motor can withstand is 0.4 N.m.

**V. Conclusion**

The proposed self-rescue mechanism enables the robot to have the following characteristics: 1- The robot can negotiate the large obstacles in pipeline that the stabilizer controller is not able to manage and the obstacle brings the arm out of region of attraction that is considered for the stabilizer controller. 2- The robot can have self-control over its outer diameter. This feature is important for the insertion and extraction of the robot from the pipeline network. The rescue mechanism is analyzed and a controller was developed based on the dynamical equations of the robot along with the rescue mechanism. The feedback for the controller is provided by the current sensors of the main motors. Since the rescue mechanism is located in dorsal side of the robot, it does not change the interference of the robot to the water flow. In our future works, we plan to perform field tests in distribution systems.

**References**

[1] S. Kazeminasab, M. Aghashahi, and M. K. Banks, "Development of an Inline Robot for Water Quality Monitoring," in 2020 5th International Conference on Robotics and Automation Engineering (ICRAE), 2020, pp. 106–113, doi: 10.1109/ICRAE50850.2020.9310805.
[2] E. Canada, "Threats to water availability in Canada." National Water Research Institute Burlington, 2004.
[3] A. L. Vickers, "The future of water conservation: Challenges ahead," J. Contemp. Water Res. Educ., vol. 114, no. 1, p. 8, 1999.
[4] D. Chatzigeorgiou, K. Youcef-Toumi and R. Ben-Mansour, "Design of a Novel In-Pipe Reliable Leak Detector," in IEEE/ASME Transactions on Mechatronics, vol. 20, no. 2, pp. 824-833, April 2015, doi: 10.1109/TMECH.2014.2308145.
[5] R. Wu et al., "Self-powered mobile sensor for in-pipe potable water quality monitoring," in Proceedings of the 17th International

Conference on Miniaturized Systems for Chemistry and Life Sciences, 2013, pp. 14–16.

[6] Y. Wu, E. Mittmann, C. Winston and K. Youcef-Toumi, "A Practical Minimalism Approach to In-pipe Robot Localization," 2019 American Control Conference (ACC), 2019, pp. 3180-3187, doi: 10.23919/ACC.2019.8814648.

[7] H. Tourajizadeh, A. Sedigh, V. Boomeri, and M. Rezaei, "Design of a new steerable in-pipe inspection robot and its robust control in presence of pipeline flow," J. Mech. Eng. Sci., vol. 14, no. 3, pp. 6993–7016, 2020. doi: 10.15282/jmes.14.3.2020.03.0548.

[8] S. Kazeminasab, A. Akbari, R. Jafari and M. K. Banks, "Design, Characterization, and Control of a Size Adaptable In-pipe Robot for Water Distribution Systems," 2021 22nd IEEE International Conference on Industrial Technology (ICIT), 2021, pp. 39-46, doi: 10.1109/ICIT46573.2021.9453583.

[9] S. Kazeminasab and M. K. Banks, "SmartCrawler: An In-pipe Robotic System with Novel Low-frequency Wireless Communication Setup in Water Distribution Systems." TechRxiv, Jan. 2021, doi: 10.36227/techrxiv.13554197.v1.

[10] S. Kazeminasab, M. Aghashahi, R. Wu and M. K. Banks, "Localization Techniques for In-pipe Robots in Water Distribution Systems," 2020 8th International Conference on Control, Mechatronics and Automation (ICCMA), 2020, pp. 6-11, doi: 10.1109/ICCMA51325.2020.9301560.

[11] I. F. Akyildiz and E. P. Stuntebeck, "Wireless underground sensor networks: Research challenges," Ad Hoc Networks, vol. 4, no. 6, pp. 669–686, 2006. doi:10.1016/j.adhoc.2006.04.003.

[12] S. Kazeminasab and M. K. Banks, "A Localization and Navigation Method for an In-Pipe Robot in Water Distribution System through Wireless Control towards Long-Distance Inspection," in IEEE Access, vol. 9, pp. 117496-117511, 2021, doi: 10.1109/ACCESS.2021.3106880.

[13] S. Kazeminasab and M. K. Banks, "Towards Long-Distance Inspection for In-pipe Robots in Water Distribution Systems with Smart Navigation," CoRR, vol. abs/2105.0, 2021, [Online]. Available: https://arxiv.org/abs/2105.05161.

[14] S. Kazeminasab, V. Janfaza, M. Razavi and M. K. Banks, "Smart Navigation for an In-pipe Robot Through Multi-phase Motion Control and Particle Filtering Method," 2021 IEEE International Conference on Electro Information Technology (EIT), 2021, pp. 342-349, doi: 10.1109/EIT51626.2021.9491887.

[15] S. Kazeminasab, R. Jafari, and M. K. Banks, "An LQR-assisted Control Algorithm for an Under-actuated In-pipe Robot in Water Distribution Systems". In Proceedings of the 36th Annual ACM Symposium on Applied Computing (SAC '21). Association for Computing Machinery, New York, NY, USA, 811–814. 2021 doi:10.1145/3412841.3442097